\documentclass[sigconf]{acmart}
\AtBeginDocument{%
  }

\usepackage{graphicx}
\usepackage{algorithm}

\usepackage{algpseudocode}

\usepackage{listings}
\lstdefinelanguage{SQL}{
    keywords={CREATE, QUERY, FROM, SELECT, WHERE, ACCUM, PRINT, SUM, VERTEX, ALTER, RETURN, ATTRIBUTE, SPACE, PRIMARY, KEY, LOAD, VALUES, ORDER, GROUP, IS, NOT, NULL, THEN, BY, LIMIT, ADD, UPDATE, *, AND, OR, SET, REPLACE, FUNCTION, RETURNS, TRIGGER, BEGIN, END, IF, AS},
    keywordstyle=\color{blue}\bfseries,
    morecomment=[l]{--},
    morecomment=[l]{//},
    morecomment=[s]{/*}{*/},
    commentstyle=\color{green}\itshape,
    morestring=[b]',
    sensitive=true,
    stringstyle=\color{red}
}

\setcopyright{acmlicensed}
\copyrightyear{2026}
\acmYear{2026}
\acmDOI{XXXXXXX.XXXXXXX}
\acmConference[ACM CAIS]{}{May 26--29, 2026}{San Jose, CA}
\begin{document}

\title{ChronoMem: Version Control and Semantic Rollback for Large Language Model Agent Memory}

\author{Yongye Su}
\affiliation{%
  \institution{Purdue University}
  \country{USA}
}
\email{su311@purdue.edu}

\author{Wujiang Xu}
\affiliation{%
  \institution{Rutgers University}
  \country{USA}
}
\email{wujiang.xu@rutgers.edu}

\author{Chaoji Zuo}
\affiliation{%
  \institution{Rutgers University}
  \country{USA}
}
\email{chaoji.zuo@rutgers.edu}

\author{Elisa Bertino}
\affiliation{%
  \institution{Purdue University}
  \country{USA}
}
\email{bertino@purdue.edu}

\newcommand{\name}{\texttt{ChronoMem}}
\newcommand{\ys}[1]{{\color{orange}\textit{#1--ys}}}

\begin{abstract}
LLM agents increasingly rely on long-term memory to support multi-session interaction and personalization. However, existing agent memory systems are designed around forward-only evolution-continuously accumulating, consolidating, and overwriting knowledge-with no principled mechanism to inspect, version, or revert prior states. This makes agents brittle under corrections, concept drift, and memory corruption, particularly after the agent has already been exposed to subsequent information. We present \name, a semantic version-control layer for agentic memory, integrated into a production-ready open-source agent framework (Agent Development Kit by Google). \name~commits whole-memory snapshots at each memory write, maintains structured version histories, and supports natural-language rollback requests by mapping ``undo'' intents to concrete historical versions via hybrid lexical–semantic retrieval, rank fusion, and reranking. We further introduce a post-exposure evaluation protocol that tests whether an agent can behave counterfactually after rollback-answering queries and summarizing history as if future updates never occurred. On long-horizon conversational benchmarks augmented with evolving memory states and rollback tasks, \name~substantially improves rollback-consistent question answering and history summarization relative to prompt-only and retrieval-only baselines, while achieving strong recovery on semantic version selection. To our knowledge, \name~is the first open-source system and benchmark for systematic, semantic, global memory rollback in LLM agents.

\end{abstract}




\maketitle

\section{Introduction}
Large language model (LLM) agents increasingly rely on \emph{long-term memory} to support multi-session conversations~\cite{locomo2024, packer2023memgpt} and personalization~\cite{li-etal-2025-hello, chhikara2025mem0}. Modern agent frameworks extract facts, summaries, and episodic traces from interactions and persist them in external memory stores or knowledge bases~\cite{zhang2024surveymemorymechanismlarge, park2023generative}. Over time, this process produces a continuously evolving memory state whose updated contents directly shape the agent's downstream outputs.
Following recent work, long-term memory in agents is typically externalized as textual traces or structured databases rather than encoded solely in model parameters, and should therefore be evaluated under incremental, multi-turn ingestion rather than in a single-block long-context setting~\cite{memoryagentbench2025}.

However, current LLM agents typically treat long-term memory as an append-only or overwrite-only store~\cite{sandbox2023}: once knowledge is written, it is difficult to inspect its evolution, revert to a previous state, or reliably ``undo'' a faulty update. This lack of version control and rollback makes agents vulnerable to error accumulation, outdated or contradictory knowledge, and memory poisoning~\cite{agentpoison2024}. Recent proprietary cloud services have begun exposing built-in APIs for per-memory history management, but these operate at the level of opaque version identifiers (e.g., rollback to a \texttt{target\_revision\_id}) and offer no public evaluation of how versioned memory affects agent behavior~\cite{vertexAI}. This gap raises several fundamental challenges:
(i) \emph{Semantic targeting:} How can an agent map a natural-language ``undo'' request to the \emph{correct} historical version without requiring users to know internal version IDs?
(ii) \emph{Global vs. local rollback:} What is the right rollback granularity---per-memory-entry versions as exposed by managed services, or \emph{whole-memory snapshots} needed to recover a temporally consistent state for downstream tasks such as global history summarization?
(iii) \emph{End-to-end consistency:} When memory is backed by external indexes/corpora and tool calls induce side effects, how can rollback restore a \emph{globally consistent} state across local memory, remote stores, and tool-induced artifacts, given that existing rewind mechanisms are typically session-scoped, and explicitly do not restore app/user-level resources nor external dependencies?
\footnote{\url{https://google.github.io/adk-docs/sessions/session/rewind/}}
\footnote{\url{https://openai.github.io/openai-agents-python/ref/run_internal/session_persistence/}}

While rollback has been explored for web navigation and version-control metaphors have been applied to \emph{context management} in other settings~\cite{zhang2025enhancing, gcc2025}, to our knowledge no \emph{open-source} system or benchmark yet \emph{systematically} evaluates \emph{semantic, global} memory rollback under a \emph{post-exposure} protocol over long-term conversational data. Existing long-horizon memory benchmarks assess memory agents along dimensions such as accurate retrieval, test-time learning, long-range understanding, and selective forgetting, but they assume a forward-evolving memory state and do not test rollback-to-a-prior-global-state behavior under post-exposure evaluation~\cite{locomo2024, memoryagentbench2025}.

\begin{figure*}[t]
    \includegraphics[width=\linewidth]{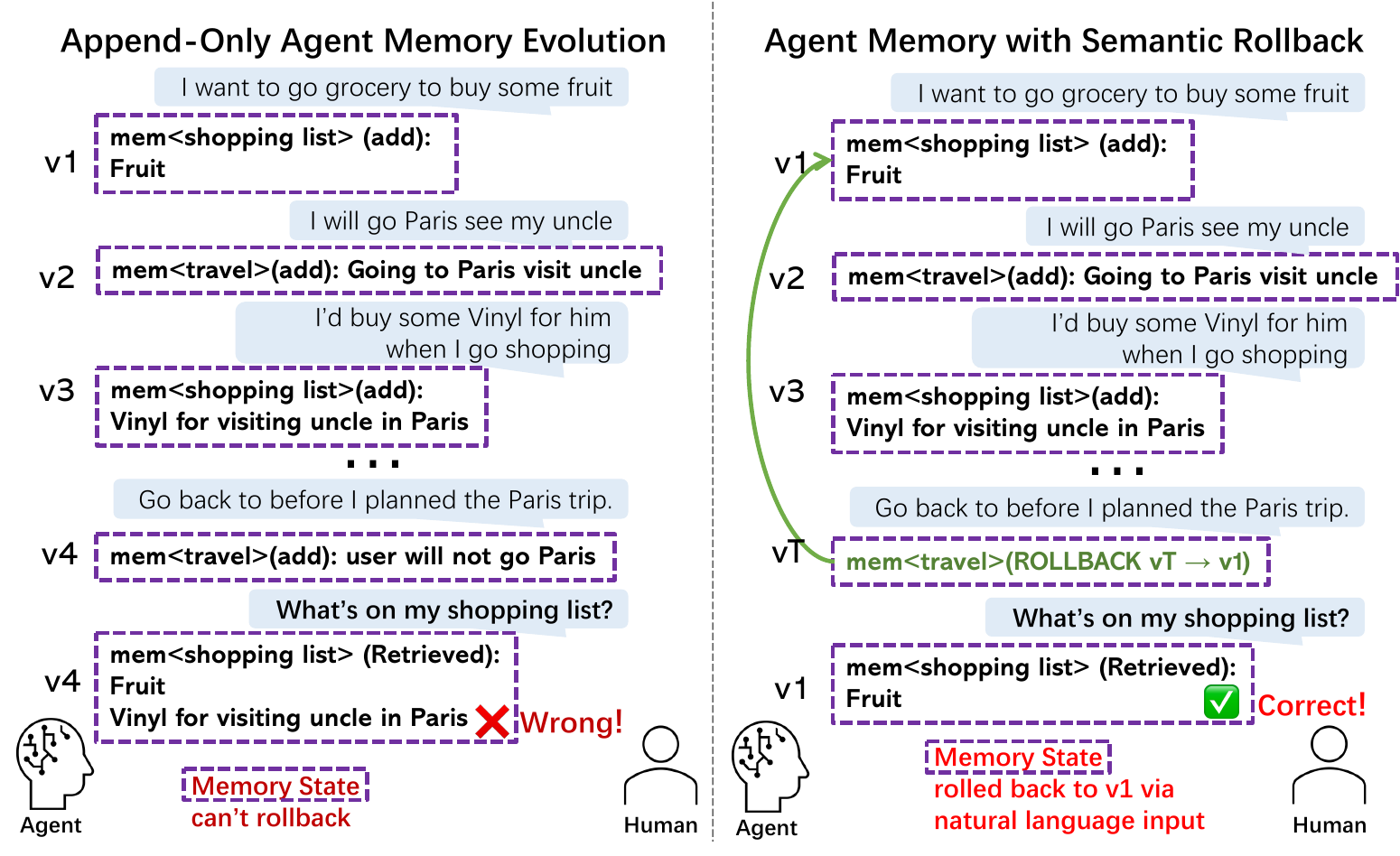}
    \caption{Comparison between append-only versus version-controlled agent memory with semantic rollback. 
    Left: Irreversible memory only moves forward-reads are bound to the latest state $S(v_T)$.
    Right: \name's memory-level semantic rollback enables a controlled memory state versioning ($v_T \rightarrow v_1$), rolling back to previous memory state $S(v_1)$ and enforcing historical consistency in subsequent reads.}
    \label{fig:memory_w_and_wo_rollback}
\end{figure*}

For example, an agent may progressively update its memory with new user preferences, facts, or plans that later prove incorrect, outdated, or even adversarially injected. In such cases, users may wish to roll back to a prior trusted memory state with a simple, natural-language ``undo'' instruction, yet mainstream agent stacks provide no principled semantic mechanism for doing so. Prior work has largely advanced agent memory in the forward direction---how to continuously accumulate, consolidate, and retrieve evolving memories from natural-language interactions. In this work, we investigate the reverse direction: after an agent has already observed subsequent interactions, can we restore a prior global memory state via semantic version control and enable counterfactual behavior---that is, ``undo'' at the memory-fact level as if the intervening interactions never occurred? Figure~\ref{fig:memory_w_and_wo_rollback} illustrates the contrast between append-only memory evolution and our version-controlled memory with semantic rollback.

We introduce \textbf{\name}, a semantic version-control layer for agentic memory, built within a production-ready and open-sourced agent framework\footnote{Agent Development Kit (ADK) by Google~\url{https://google.github.io/adk-docs/}}. \name~ equips LLM agents with time-aware memory snapshots, version histories, and rollback operations that can be driven either by explicit version identifiers or by natural-language descriptions of the desired prior state.

Our contributions are as follows:
\begin{itemize}
    \item We design and implement a \emph{version control abstraction for LLM agent memory} within an open-source production-ready framework (ADK by Google), introducing global snapshots, version histories, and commit semantics for agent memory, and integrating these with a retrieval-backed persistent memory index backend.

    \item We propose the \emph{first open-source semantic global rollback mechanism for agent memory} that maps natural-language ``undo'' requests to whole-memory versions, and implement an end-to-end rollback pipeline with verifiable integrity checks.

    \item We introduce a systematic evaluation standard for \emph{post-exposure semantic rollback} of agent memory by augmenting long-term conversational memory benchmarks with evolving memory states, updates, and rollback tasks.
\end{itemize}

\name~ establishes a new foundation for controllable long-term memory in LLM agents, moving from passive accumulation to actively managed, reversible, and semantically searchable memory states.

\begin{table*}[t]
\centering
\small
\setlength{\tabcolsep}{5pt}
\begin{tabular}{lccccc}
\toprule
\textbf{Work} & \textbf{Rollback Target} & \textbf{Targeting Interface} & \textbf{Open-source-based} & \textbf{Public Rollback Eval} \\
\midrule
\textbf{\name} (ours) & \textbf{Global Agent Memory} & \textbf{Natural Language $\rightarrow$ Rollback} & \checkmark & \checkmark \\
\midrule

Vertex AI Memory Revisions & Memory-id specific revision & Revision ID (\texttt{target\_revision\_id}) & $\times$ & $\times$ \\
ADK Session Rewind & Session state/artifacts & Invocation/rewind point ID & \checkmark & $\times$ \\
OpenAI Agents Sessions & Session history items & \texttt{pop\_item}/history ops & \checkmark & $\times$ \\
WebRollback & Navigation trajectory & Explicit rollback action & $\times$ & \checkmark \\
\bottomrule
\end{tabular}
\vspace{1mm}
\caption{Scope comparison of rollback mechanisms. Vertex provides per-memory revision history and rollback via a revision identifier (e.g., \texttt{target\_revision\_id})~\cite{vertexAI}. ADK rewind is session-scoped and explicitly does not restore app/user-level resources nor external dependencies~\cite{adkRewind}. OpenAI Agents SDK Sessions manage conversation history with operations such as \texttt{pop\_item} for undoing recent turns and are released under an open-source license~\cite{openaiSessions}. WebRollback evaluates rollback over navigation trajectories~\cite{zhang2025enhancing}. Existing long-horizon memory benchmarks (e.g., LoCoMo, MemoryAgentBench) do not test rollback-to-a-prior-global-memory-state behavior~\cite{memoryagentbench2025,locomo2024}.}
\label{tab:scope}
\end{table*}

\section{Related Work} 

\subsection{LLM Agent Memory}
Long-term memory has become essential for LLM agents operating across multi-session interactions. Early systems introduced hierarchical memory management, reflection and summarization over stored observations, and forgetting-curve-based update mechanisms to enable agents to maintain context beyond fixed context windows~\cite{park2023generative,packer2023memgpt,zhong2024memorybank}. A-Mem~\cite{amem2025} shifted from passive storage to agentic memory, where the agent itself dynamically organizes and links memories through structured note-linking rather than relying on fixed retrieval operations, marking a transition toward treating memory as a first-class, self-organizing structure. Building on these foundations, more recent works~\cite{chhikara2025mem0,rasmussen2025zep} have scaled toward production deployment, introducing scalable memory extraction and consolidation pipelines, graph-based representations that capture relational structures among conversational elements, and temporal knowledge graphs that maintain bi-temporal models tracking both when facts occurred and when they were ingested. These systems have been evaluated on increasingly comprehensive long-term conversational benchmarks that test factual recall, temporal reasoning, and multi-hop inference across hundreds of dialogue turns spanning dozens of sessions~\cite{locomo2024}. Despite this progress, existing systems primarily focus on memory evolution (adding, updating, consolidating, and selective forgetting) and typically do not expose explicit version-control primitives that preserve whole-memory snapshots or enable rollback to a prior global memory state. Once a memory is overwritten or merged, prior states are often not retained in an accessible, auditable form, leaving agents vulnerable to error accumulation and memory poisoning attacks where adversarially injected entries can propagate through future retrievals~\cite{agentpoison2024}. To our knowledge, no existing open-source system offers whole-memory snapshots, commit-level versioning, or natural language driven semantic rollback integrated into a production-ready agent framework, which are the gaps that \name~is designed to fill.

\subsection{Traceable Version Control and Rollback}
Rollback operations rely on traceable version control and persistent logging. In LLM agent memory, unlike selective forgetting (e.g., revising beliefs to resolve contradictions), rollback is a \emph{reversible state-control} primitive that restores a prior memory state and enables counterfactual behavior after the agent has already been exposed to later information.

Several works study how agent memories are edited or adapted over time. Memory Sandbox exposes conversational memory as editable objects that users can inspect, modify, or delete, improving transparency but relying on manual operations~\cite{sandbox2023}. Dynamic recall models introduce human-like decay and consolidation so that frequently used facts remain accessible and obsolete ones fade~\cite{dynamicRecall2024}. These approaches manage \emph{how} memory changes, but once an update is applied there is no automatic way to restore a prior global snapshot.

Closer to our setting, there is emerging work on rollback for other forms of LLM-driven state. Git-Context Controller (GCC) treats an agent’s code context like a version-controlled filesystem, with commit, branch, and merge operations for reasoning traces~\cite{gcc2025}. Database research proposes a middleware layer that treats LLM-generated database transactions as removable and uses invariant-based coordination so that rolling back a transaction preserves ACID consistency~\cite{zeng2024simplefastwayhandle}. In WebRollback~\cite{zhang2025enhancing}, web agents can revert to previous navigation states in a search tree and re-plan, improving robustness on complex web tasks. In industry practice, Vertex~AI Agent Engine added a proprietary Memory Revisions feature that records individual memory edits and supports rollback via a revision (version) identifier (e.g.,\texttt{target\_revision\_id}) in a managed memory bank~\cite{vertexAI}. In addition, common framework utilities such as ADK session rewind and OpenAI Agents SDK session persistence can undo session-scoped history items, but do not provide semantic rollback over globally versioned long-term memory states~\cite{adkRewind,openaiSessionPersistence,openaiSessions}. Table~\ref{tab:scope} shows the aforementioned scope difference. Finally, AgentPoison shows that poisoning an agent’s memory or knowledge base can induce persistent backdoor behaviors, underscoring the need for stronger control over what agents remember~\cite{agentpoison2024}. Compared to these efforts, \name~targets internal long-term agent memory rather than database or environment state, is implemented in an open-source agent framework, and focuses specifically on \emph{semantic} version selection and \emph{global} rollback to whole-memory snapshots rather than ID-only, per-memory-entry rollback or manual rewind.

\subsection{Memory Retrieval and Evaluation} 
Work on long-term memory evaluation emphasizes retrieval and usage, not version control. LoCoMo proposes a benchmark of very long, multi-session conversations and evaluates whether agents can recall and use information from earlier sessions for question answering and event summarization~\cite{locomo2024}. MemoryAgentBench defines four competencies for memory-enabled agents-accurate retrieval, test-time learning, long-range understanding, and selective forgetting-and measures performance on incremental multi-turn tasks~\cite{memoryagentbench2025}. These benchmarks reveal that current systems struggle with long-horizon memory and forgetting, even when equipped with vector databases or extended context windows. However, they assume a single evolving memory state and do not consider choosing among \emph{multiple} historical memory versions or restoring a prior global memory state under post-exposure rollback.

In contrast to prior work, \name~ provides a framework-level, open-source mechanism for \emph{semantic version control and rollback of agent memory}. It maintains a versioned chronological memory within ADK, maps natural-language “undo” requests to concrete memory snapshots, and consistently maintains both local memory and retrieval-backed memory indexes. Our evaluation augments long-horizon conversational memory benchmarks with evolving memory states and rollback probes, and measures rollback-consistent QA/summarization under post-exposure, providing the first systematic evaluation of semantic global rollback for versioned agent memory in long-term conversational settings.


\section{System Design of ChronoMem: The Natural Language ``Undo'' Button of Agent Memory}
\label{sec:system_design}

\subsection{Goal and Scope}
\name\ provides \emph{version-controlled, reversible long-term memory} for LLM agents running in a modern agent framework.
We target agents built on Google’s open-sourced Agent Development Kit (ADK), where an agent execution is orchestrated by a runtime event loop (Runner) and persists interaction traces as session events and state, while optionally using a \texttt{MemoryService} to maintain searchable long-term knowledge across sessions.
While ADK supports session-level rewind for reverting a session to a previous request state, rewind is explicitly scoped to session-managed resources and does not restore app/user-level state or external dependencies~\cite{adkRewind}.

Our goal is to enable \emph{global memory version control} for LLM-based agent: every memory write produces a new \emph{whole-memory snapshot} (a commit), and rollback restores the \emph{entire} long-term memory state to a prior snapshot.
This capability is critical for \emph{post-exposure} settings: after an agent has already observed later interactions, we must be able to restore a historical memory state such that downstream behavior (QA and summarization) is \emph{counterfactual}---as if post-$v^{*}$ information never occurred.

\name\ focuses on (i) defining a systems-level abstraction for global memory commits and rollback at the agent memory boundary, and (ii) implementing a semantic control plane that maps natural-language ``undo'' requests to concrete historical versions.
We do not attempt to undo arbitrary external side effects (e.g., third-party services) beyond the agent memory substrate; consistent restoration of external systems remains application-dependent, consistent with the limitations of existing rewind mechanisms~\cite{adkRewind}.

\subsection{Architecture Overview}
\begin{figure*}
    \centering
    \includegraphics[width=\linewidth]{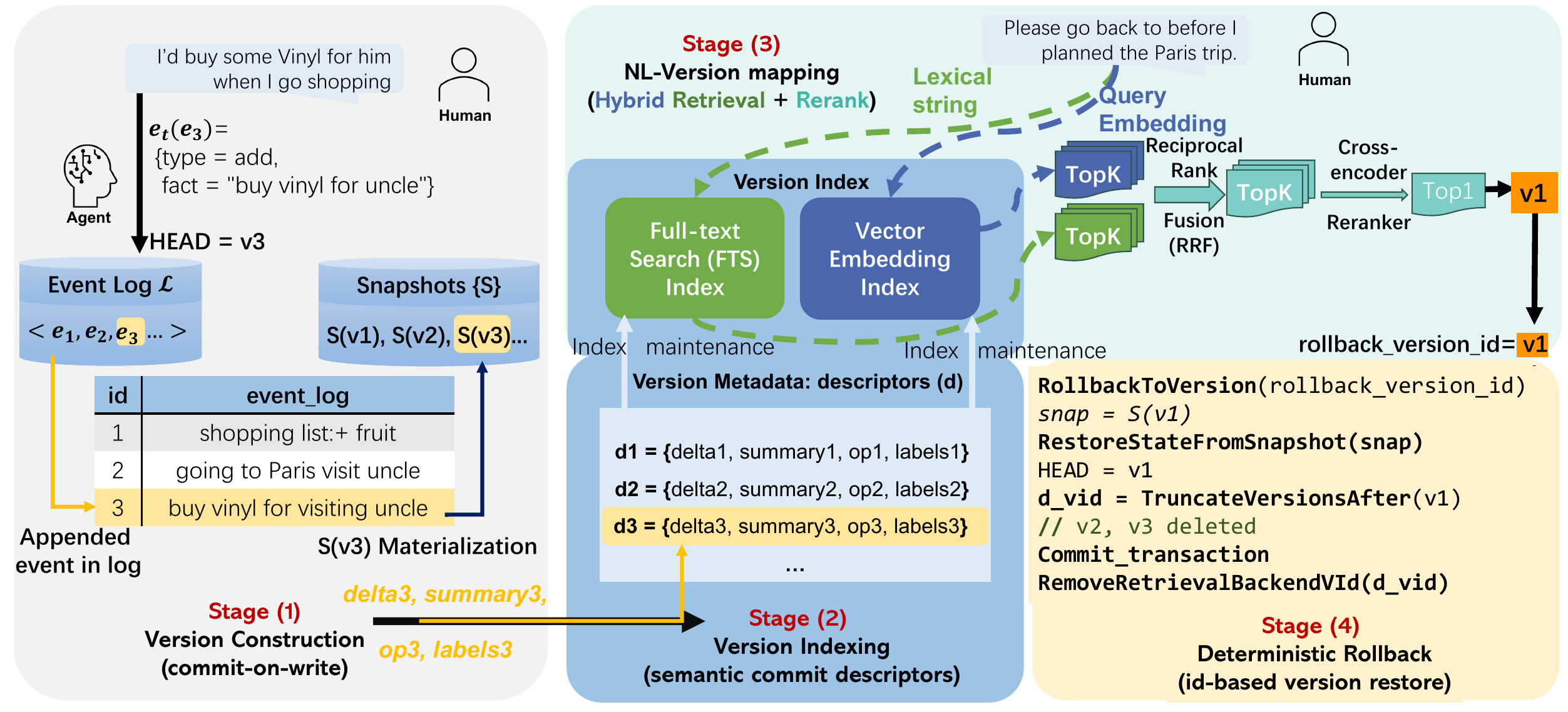}
    \caption{The system architecture}
    \label{fig:arc}
\end{figure*}

The Figure~\ref{fig:arc} depicts the rollback path from the data layer, index layer, to the control layer.
\name\ is implemented as a drop-in, versioned \texttt{MemoryService} wrapper and consists of:

\begin{itemize}
  \item \textbf{LocalMemoryService (orchestrator).} Intercepts memory writes to create versions and persists them; exposes rollback APIs such as 
  \begin{description}
    \item[\texttt{rollback\_to\_version(version\_id)}] deterministically \\ restores the snapshot $S(v_{vid})$ and sets \texttt{HEAD} := $vid$. \item[\texttt{rollback\_to\_version\_by\_nl\_query(query)}] resolves \\ a natural-language rollback request $q$ to a target version $\hat{v}$ via hybrid retrieval over semantic commit descriptors, then invokes \texttt{rollback\_to\_version($\hat{v}$)}.
  
  \end{description}
  Both APIs are atomic at the per-(app,user) scope for identification purposes and preserve the interface invariant that all subsequent memory reads are scoped to the active version collections.
  
  \item \textbf{SQLite memory store.} Persists the append-only event log, version metadata, and snapshot payloads, and maintains the active version pointer (\texttt{HEAD}).
  \item \textbf{Reranker module.} Refines candidate rollback versions using a stronger relevance model (e.g., a cross-encoder reranker).
\end{itemize}

\paragraph{Interface invariant.}
ChronoMem enforces a simple invariant at the memory boundary: reads are scoped to the active version (\texttt{HEAD}).
After rollback to $v^*$, any subsequent memory read or retrieval must behave as if only the state at $v^*$ exists (no post-$v^*$ visibility).

\subsection{Version Control and Semantic Rollback}
\label{sec:vc_rollback}

ChronoMem exposes agent memory rollback as a systems problem: given a natural-language undo request $q$,
select a target historical \emph{version} $\hat{v}$ and restore the agent's long-term memory state so that
all subsequent memory reads are scoped to that restored state. This enables \emph{post-exposure} counterfactual behavior:
the agent may have observed interactions after the desired point, but after rollback it must behave as if only the
rolled-back memory were available.

\paragraph{State model: versions, snapshots, and the event log.}
We model long-term agent memory as an append-only event log $\mathcal{L}$ with explicit versions $\{v_j\}$ and snapshots $S(v_j)$, notably, we use $v_j$ and \texttt{version\_id} interchangeably to denote a globally unique version identifier.
Each memory write generates an immutable event $e_i$ appended to a serial log:
\[
\mathcal{L}=\langle e_1,e_2,\ldots e_j \rangle,
\]
following the event-sourcing principle that historical states can be reconstructed from the event stream~\cite{fowlerEventSourcing}.
For auditability and recovery, this mirrors write-ahead logging ideas (e.g., ARIES) in spirit, but operates at the level of logical agent-memory state rather than page-level database recovery~\cite{mohan1992aries}.
Each event $e_i$ is associated with a record timestamp $\tau_i$ indicating when the update is recorded.

A version $v_j$ corresponds to a whole-memory snapshot materialized from a prefix of the log:
\[
S(v_j) \triangleq \mathrm{snap}(\langle e_1{...}e_{t_j}\rangle), \qquad where~ t_1<t_2<\cdots<t_j
\]
where $\mathrm{snap}(\cdot)$ serializes the full agent-memory state required by downstream reasoning and ${t_j}$ is the maximal event sequence number associated with version.
This log+snapshot design avoids replaying the entire log during rollback and trades storage for fast restoration.

\paragraph{\textbf{Stage (1) Version construction: commit-on-write snapshots.}}
Our work \name~ creates a new version after each memory write (commit-on-write) to support fine-grained rollback targets.
Operationally, a commit appends the new event to $\mathcal{L}$, materializes $S(v_j)$, stores metadata, and advances the active pointer \texttt{HEAD} to the latest version.
Algorithm~\ref{alg:commit} shows the transactional write path.

\begin{algorithm}[t]
\caption{\textsc{CommitOnWrite}$(app,user,e)$}
\label{alg:commit}
\begin{algorithmic}[1]
\State \textbf{// Append event and create a new version $v_j$ (commit-on-write).}
\State {// Example: user says ``I moved to NYC'' $\Rightarrow$ append event $e$ and commit corresponding new version.}
\State \textbf{begin transaction}
\State $seq \gets$ \textsc{NextEventSeq}$(app,user)$
\State $\tau \gets$ \textsc{Now}(); \Comment{record timestamp}
\State \textsc{AppendEvent}$(app,user,seq,\tau,e)$ \Comment{immutable event log insert}
\State $vid \gets$ \textsc{NextVersionId}$(app,user)$
\State $\mathbf{d} \gets$ \textsc{BuildCommitDescriptor}$(app,user,seq)$ 
      \Comment{$\{\texttt{delta},\texttt{summary},\texttt{op},\texttt{labels}\}$}
\State $snap \gets$ \textsc{MaterializeSnapshot}$(app,user,seq)$
      \Comment{serialize whole-memory state (e.g., session\_events)}
\State \textsc{PersistVersion}$(app,user,vid,seq,\tau,\mathbf{d},snap)$ \Comment{store metadata + snapshot payload}
\State \textsc{SetHead}$(app,user,vid)$ \Comment{advance active version pointer}
\State \textbf{commit transaction}
\State \Return $vid$
\end{algorithmic}
\end{algorithm}

\paragraph{\textbf{Stage (2) Version index: semantic commit descriptors.}} Version indexing is version-level secondary index construction over semantic commit descriptors. To support natural-language rollback, each version $v_j$ is indexed by a semantic commit descriptor $\mathbf{d}_j$
that summarizes the incremental change introduced at $v_j$:
\[
\mathbf{d}_j \triangleq \{\texttt{delta}_j,\texttt{summary}_j,\texttt{op}_j,\texttt{labels}_j\}.
\]
Here $\texttt{delta}_j$ captures operation-level changes (e.g., added/updated/removed facts),
$\texttt{summary}_j$ provides a compact natural-language description,
$\texttt{op}_j$ denotes update types, and
$\texttt{labels}_j$ contains temporal metadata (e.g., session index and record timestamp $\tau$).
These descriptors form the \emph{retrieval surface} for semantic rollback.

Importantly, hybrid retrieval is performed over $\{\mathbf{d}_j\}$ (the version index), rather than over full snapshots $S(v_j)$ or raw events $\mathcal{L}$.
This ensures that rollback selection is version-level and computationally bounded by the number of commits, independent of snapshot size. 

Formally, lexical retrieval over commit descriptors scales with the number of versions $|\{v_j\}|$ under inverted indexing, while dense retrieval follows vector's approximate nearest neighbor (ANN) search complexity bounds~\cite{vexless}.
Thus, we search a smaller-order data change log rather than the entire underlying data; the selection cost is independent of snapshot size $|S(v_j)|$, since snapshot payloads are accessed only after a target version is chosen.

\paragraph{\textbf{Stage (3): NL$\rightarrow$version mapping via hybrid retrieval.}}
Given a natural-language rollback request $q$, ChronoMem selects a target version using a standard two-stage retrieval stack:
\textbf{lexical} retrieval and \textbf{dense} semantic retrieval in parallel over $\{\mathbf{d}_j\}$, followed by fusion and reranking.

\noindent\textbf{Lexical retrieval.} We use SQLite FTS5 BM25-style ranking over commit descriptors, e.g.:
\begin{lstlisting}[language=SQL, caption={Lexical retrieval over commit descriptors using SQLite FTS5 BM25 (lower bm25() is better).}, label={lst:sql_bm25}]
SELECT version_id, bm25(commit_desc_fts) AS bm25_score
FROM commit_desc_fts
WHERE commit_desc_fts MATCH 'before Paris trip conversation' 
ORDER BY bm25(commit_desc_fts) ASC, rowid ASC
LIMIT ?;
\end{lstlisting}

\noindent\textbf{Dense retrieval.} In parallel, we embed commit descriptors and retrieve top-$K$ by vector similarity.

\noindent\textbf{Fusion (RRF).} We fuse the two ranked lists with Reciprocal Rank Fusion:
\[
\mathrm{RRF}(v_j\mid q)=\sum_{m\in\{\mathrm{lex},\mathrm{sem}\}} \frac{1}{k_0+\mathrm{rank}_m(v_j)},
\]
where $\mathrm{rank}_m(\cdot)$ starts from 1 and $k_0$ is a constant (commonly set to 60 in practice).

\noindent\textbf{Reranking.} Finally, ChronoMem optionally reranks the top-$K$ fused candidates with a stronger relevance model
(e.g., a cross-encoder) and selects $\hat{v}$.

\begin{algorithm}[t]
\caption{\textsc{RollbackToVersion}$(app,user,vid)$}
\label{alg:rbv}
\begin{algorithmic}[1]
\State \textbf{begin transaction}
\State $snap \gets$ \textsc{LoadSnapshot}$(app,user,vid)$
\State \textsc{RestoreStateFromSnapshot}$(app,user,snap)$
\State \textsc{SetHead}$(app,user,vid)$
\State \textsc{TruncateVersionsAfter}$(app,user,vid)$
\State \textbf{commit transaction}
\State \textsc{SynchronizeRetrievalBackendIfAny}$(app,user,vid,snap)$ \Comment{best-effort; idempotent}
\State \Return $vid$
\end{algorithmic}
\end{algorithm}

\paragraph{\textbf{Stage (4): Version rollback execution (ID-based).}}
Semantic rollback is a control-plane operation that resolves $q$ to a concrete version identifier $\hat{v}$.
The actual state transition is executed by a deterministic data-plane primitive:
\[
\textsc{RollbackToVersion}(app,user,\hat{v}),
\]
which restores the snapshot $S(\hat{v})$ and updates \texttt{HEAD}.
Algorithm~\ref{alg:rbv} shows the rollback path: load snapshot payload, restore the local memory state, set \texttt{HEAD}, and truncate later versions to maintain a linear history.
If a retrieval-backed index is used, we synchronize it best-effort after the local state commit.

As a last resort, \name~ maintains an explicit ID-based API (\texttt{rollback\_to\_version(version\_id)}) as a primitive for version maintenance and reproducibility.
Semantic rollback function interface (\texttt{rollback\_to\_version\_by\_nl\_query(query)}) composes over it:
it resolves natural language query mapping $q\mapsto \hat{v}$ via retrieval and then invokes the same deterministic restore path.
This separation avoids conflating (i) selection errors with (ii) restoring correctness, and makes evaluation interpretable.

\section{Experiment Setup} 
\label{sec:experiment_setup}
We evaluate \name\ on two long-horizon memory-agent benchmarks and adapt them to a rollback setting, focusing on two downstream tasks: question answering and event summarization. Our experiments are designed to answer the following questions: 

\begin{description}
\item[\textbf{(RQ1)} \emph{Semantic version selection}:] \leavevmode\\
Given a natural-language rollback request, how accurately does \name~ select the intended target version $v^{*}$ among historical snapshots? This is to prevent conflating task performance with the correctness of version selection.

\item[\textbf{(RQ2)} \emph{Rollback-consistent QA}:] \leavevmode\\
After rolling back to a target memory version $v^{*}$, can the agent answer questions correctly \emph{as if only $v^{*}$ were available}, without leaking post-$v^{*}$ information?

\item[\textbf{(RQ3)} \emph{Rollback-consistent summarization}:] \leavevmode\\
After rolling back to $v^{*}$, can the agent generate event summaries that are temporally consistent with the historical state at $v^{*}$, rather than reflecting later updates?

\end{description}


\subsection{Datasets and Usage}
\paragraph{LoCoMo} It is a very long-term conversational benchmark with multi-session dialogues and tasks, including question answering and event summarization~\cite{locomo2024}. We use LoCoMo to evaluate rollback-consistent QA and rollback-consistent history summarization.

\paragraph{MemoryAgentBench (MAB)} It evaluates memory agents under incremental multi-turn ingestion and formalizes competencies including Accurate Retrieval, Test-Time Learning, Long-Range Understanding, and Conflict Resolution~\cite{memoryagentbench2025}. We use MAB to stress semantic rollback under frequent updates.

\paragraph{Post-exposure rollback protocol.}
Both datasets are originally designed for forward-evolving memory. To evaluate \emph{post-exposure} rollback, we run the agent according to two phases:
(1) \textbf{Exposure:} ingest the full interaction stream to the final state $v_T$; 
(2) \textbf{Rollback:} issue a natural-language rollback request and restore to a specific version $\hat{v}$; then execute the downstream task (QA or summarization) on the restored state.
This protocol enforces a counterfactual requirement: the agent has already observed interactions after $v^{*}$, but should behave as if only the rolled-back state were available. Here we define the ground-truth target $v^{*}$ by the annotated evidence scope (LoCoMo) ~\cite{locomo2024} and by the constructed update boundary in incremental streams (MAB)~\cite{memoryagentbench2025}. 

\paragraph{Rollback query construction.}
To evaluate semantic rollback, we construct natural-language rollback requests that specify an intended historical state $v^{*}$ without exposing internal version identifiers.
For each instance, we first derive a \emph{grounded anchor} from dataset annotations available at or before $v^{*}$.
For LoCoMo, we use the annotated QA evidence spans (dialog IDs containing the answer) and session-level event annotations as anchors when available~\cite{locomo2024}.
For MAB, we use the constructed update boundary and the associated pre-update memory content as the anchor~\cite{memoryagentbench2025}.
We then generate rollback queries in two ways: (i) \emph{ground-truth-backed templates} (e.g., ``undo the update about \{anchor\}''; ``go back to when \{anchor\} was true''), and (ii) \emph{LLM-generated paraphrases} of these templates to increase linguistic diversity.
To avoid trivializing version selection, we disallow explicit version IDs, timestamps, or session indices in rollback queries. We constrain LLM generation to preserve the anchor semantics and discard any query whose reference does not uniquely identify the intended historical scope.

\subsection{Metrics}

We report both version selection quality (RQ1) and downstream correctness after semantic rollback (RQ2--RQ3) as a result of semantic rollback.

\paragraph{Semantic version selection (RQ1).}
Let $\hat{v}$ be the selected version and $v^{*}$ the ground-truth target rollback version. We evaluate:
\begin{itemize}
    \item \textbf{Recall@1:} whether $\hat{v} = v^{*}$.
    \item \textbf{Recall@$k$:} whether $v^{*}$ appears in the top-$k$ retrieved candidates before final selection.
    \item \textbf{Scope@$\frac{k}{2}$:} temporal locality defined as $\Pr[|\Delta|\le \frac{k}{2}]$, where $\Delta = |\hat{v} - v^{*}|$ in target rollback index (or session index when aligned).
\end{itemize}
Scope@$\frac{k}{2}$ diagnoses ``near-miss'' behavior when adjacent versions are semantically similar.
While Recall@$k$ measures whether the ground-truth version $v^*$ appears among the top-$k$ retrieved candidates, 
Scope@$\frac{k}{2}$ instead evaluates temporal proximity by measuring whether the selected version $\hat{v}$ lies within $\frac{k}{2}$ steps of $v^*$ along the version timeline (i.e., $|\hat{v}-v^*|\le k$).
Thus, although both metrics involve a parameter $k$, Recall@$k$ reflects ranking accuracy within the candidate set, whereas Scope@$\frac{k}{2}$ reflects temporal locality of the final 1 selection.

\paragraph{Rollback-consistent QA (RQ2).}
We evaluate QA outputs produced \emph{after rollback} using dataset ground-truth answers, reporting Exact Match and/or F1 following prior benchmark practice~\cite{locomo2024}. When applicable, we additionally report whether the answer contains post-$v^{*}$ facts, treating any reliance on later information as a rollback-consistency violation.

\paragraph{Rollback-consistent summarization (RQ3).}
For event/history summarization, we compute ROUGE-1 against reference summaries corresponding to the target historical scope~\cite{lin2004rouge}. Higher ROUGE indicates that the restored memory state yields summaries consistent with the intended timeline.

\subsection{Baselines}
We compare \name\ (semantic global rollback) against baselines that approximate rollback without restoring a global memory snapshot.
\paragraph{Prompt-only rollback.}
We provide the natural-language rollback instruction together with the downstream task query, while keeping the underlying memory state unchanged. 
This baseline evaluates whether the LLM can implicitly infer the intended historical state from the conversation and suppress post-$v^{*}$ information through instruction following alone, without structural state isolation.

\paragraph{Full-history prompt rollback.}
We include the entire interaction history in the model input (including content after $v^{*}$) and instruct the model to answer ``as if'' rolled back. 
This represents a stronger post-exposure stress test under maximal contamination, where the model must reason over the full context and ignore information beyond the intended rollback boundary.

\paragraph{Retrieval-only rollback (vector-db only).}
We approximate rollback by retrieving relevant historical memory items using dense similarity search (ANN top-$k$) and generating answers conditioned on the retrieved snippets, without restoring a versioned snapshot. 
This baseline isolates the effect of explicit version-scoped state restoration from standard retrieval augmentation.

\section{Evaluation}

This section presents a comprehensive empirical evaluation of ChronoMem under the rollback protocol defined in Section~\ref{sec:experiment_setup}. Our goal is not only to measure task performance but to isolate and quantify the architectural contribution of explicit version restoration. Unlike standard memory-agent evaluations that assume forward-only evolution, our setting introduces a structural state transition prior to task execution; we therefore evaluate ChronoMem along two orthogonal axes: (i)~the correctness of semantic version resolution, and (ii)~the behavioral correctness of the agent after state restoration. All downstream metrics are reported on the same set of rollback instances to ensure that improvements in QA or summarization are not confounded with version-selection quality.

Our evaluation explicitly tests whether rollback consistency can be achieved solely through reasoning or whether it requires architectural support for historical state management. By comparing ChronoMem against prompt-only and retrieval-based baselines with comparable backbone models, we aim to demonstrate that counterfactual correctness under post-exposure conditions cannot be reliably obtained without explicit version-scoped memory restoration. Unless otherwise specified, all experiments follow the exposure--rollback--execution protocol described in Section~\ref{sec:experiment_setup}. We report results on both LoCoMo~\cite{locomo2024} and MemoryAgentBench~\cite{memoryagentbench2025}, covering semantic version selection accuracy, rollback-consistent QA, and rollback-consistent summarization, and further analyze ablations and system-level overhead to characterize the trade-offs introduced by semantic version control.

\subsection{RQ1: Semantic Version Selection}
We first evaluate semantic version selection independently of downstream QA and summarization tasks. In this setting, no generative decoding is involved; the evaluation concerns only the natural-language-to-version resolution pipeline described in Section~\ref{sec:vc_rollback}. Given a rollback query, we assess the quality of version targeting using hybrid retrieval followed by cross-encoder reranking. Unless otherwise specified, we use Cohere Rerank-3.5 as the cross-encoder model throughout.\footnote{\url{https://cohere.com/blog/rerank-3pt5}} All metrics are computed over the selected version identifiers without invoking the backbone LLM used for task execution.

As shown in Tables~\ref{tab:rq1_locomo} and~\ref{tab:rq1_mab}, \name\ substantially outperforms all baselines in Recall@1 and Recall@5 on both datasets, demonstrating that rollback version selection benefits from explicit version-level indexing and hybrid retrieval rather than from prompt-based self-suppression or snippet-level similarity search. On LoCoMo, \name\ improves Recall@1 from 12.0\% (Hybrid) to 20.5\% and Recall@5 from 28.1\% to 38.9\%; similar gains are observed on MAB (24.3\%$\rightarrow$39.4\% in Recall@1, 53.8\%$\rightarrow$65.2\% in Recall@5). The consistent improvements across both benchmarks indicate that explicit version-level indexing combined with temporal-aware reranking is critical for reliable rollback selection. Scope@2 further reveals that even when exact matching fails, \name\ more frequently selects temporally adjacent versions (e.g., 31.2\% vs.\ 17.9\% on LoCoMo), suggesting that temporal-aware reranking effectively disambiguates semantically similar neighboring versions. Overall, hybrid fusion alone is insufficient; the reranking stage is essential for resolving such cases. These results validate our system design and confirm that semantic rollback is fundamentally a version-selection problem over structured historical states rather than a conventional snippet-level retrieval task.

\begin{table}[t]
\centering
\small
\begin{tabular}{l|ccc}
\toprule
\textbf{Method} & \textbf{Recall@1} $\uparrow$ & \textbf{Recall@5} $\uparrow$ & \textbf{Scope@2} $\uparrow$ \\
\midrule
BM25-only & 9.3\%  & 22.3\% & 13.1\% \\
Dense-only & 8.7\% & 23.6\% & 12.8\% \\
Hybrid (RRF)  & 12.0\% & 28.1\% & 17.9\% \\
\name & \textbf{20.5\%} & \textbf{38.9\%} & \textbf{31.2\%} \\
\bottomrule
\end{tabular}

\caption{Semantic version selection performance on LoCoMo.}
\label{tab:rq1_locomo}
\end{table}

\begin{table}[]
    \centering
    \begin{tabular}{l|ccc}
    \toprule
    \textbf{Method} & \textbf{Recall@1} $\uparrow$ & \textbf{Recall@5} $\uparrow$ & \textbf{Scope@2} $\uparrow$ \\
    \midrule
    BM25-only & 14.1\%  & 44.6\% & 23.3\% \\
    Dense-only & 12.7\% & 46.4\% & 24.9\% \\
    Hybrid (RRF)  & 24.3\% & 53.8\% & 40.5\% \\
    \name & \textbf{33.4\%} & \textbf{60.2\%} & \textbf{58.0\%} \\
    \bottomrule
    \end{tabular}
    \caption{Semantic version selection performance on MAB }
    \label{tab:rq1_mab}
\end{table}


\subsection{RQ2: Rollback-Consistent QA}
Tables~\ref{tab:rq2_locomo} and~\ref{tab:rq2_mab} report question-answering performance \emph{after} semantic rollback under the post-exposure protocol. In contrast to prompt-only baselines, which attempt to suppress post-$v^*$ information through instruction following, and retrieval-only approximations, which rely on snippet-level similarity search without restoring a global memory state, \name\ consistently achieves the best post-rollback QA performance across all three backbone models. On LoCoMo, \name\ improves QA F1 over the strongest non-versioned baseline (RAG-only) for all backbones, indicating that explicit version-scoped state restoration provides benefits beyond standard retrieval augmentation on long-horizon conversations~\cite{locomo2024}. Notably, prompt-only rollback performs near chance, suggesting that post-exposure contamination is difficult to overcome through prompting alone once the agent has already observed subsequent interactions. On MAB, where we restrict evaluation to the Accurate Retrieval category due to task-scope constraints, \name\ again yields the strongest performance across all backbones, confirming that deterministic global snapshot restoration is effective for counterfactual QA in an incremental, multi-turn memory setting.



\begin{table}[h]
\centering
\small
\begin{tabular}{l|ccc}
\toprule
\textbf{Method} 
& \textbf{Llama-3.1-8B} 
& \textbf{Qwen2.5-7B} 
& \textbf{Mistral-7B} \\
\midrule
Prompt-only & 2.3\% & 2.8\% & 0.9\%  \\
Full-history & 19.3\% & 20.6\% & 13.5\%   \\
RAG-only & 28.9\% & 27.1\% & 19.4\%  \\
\name & \textbf{36.1\%}  & \textbf{38.5\%} & \textbf{31.3\%} \\
\bottomrule
\end{tabular}
\caption{Rollback-consistent QA F1 (\%) across different base models on LoCoMo dataset.}
\label{tab:rq2_locomo}
\end{table}

\begin{table}[h]
\centering
\small
\begin{tabular}{l|ccc}
\toprule
\textbf{Method} 
& \textbf{Llama-3.1-8B} 
& \textbf{Qwen2.5-7B} 
& \textbf{Mistral-7B} \\
\midrule
Prompt-only & 0.8\% & 1.2\% & 0.0\%   \\
Full-history & 21.3\% & 24.7\% & 14.9\%   \\
RAG-only & 35.5\% & 34.2\% & 31.0\%   \\
\name & \textbf{53.8\%} & \textbf{55.1\%} & \textbf{44.6\%}  \\
\bottomrule
\end{tabular}
\caption{Rollback-consistent QA on MemoryAgentBench (Accurate Retrieval subset), reported as task-specific accuracy (\%)~\cite{memoryagentbench2025}.}
\label{tab:rq2_mab}
\end{table}


\subsection{RQ3: Rollback-Consistent Summarization}
Tables~\ref{tab:rq3_locomo} and~\ref{tab:rq3_mab} report summarization quality after semantic rollback on LoCoMo and MAB, respectively. Across all backbone models, \name\ achieves the best performance, indicating that explicit global snapshot restoration yields summaries that are more temporally consistent with the intended historical state. On LoCoMo, \name\ improves ROUGE-1 over the strongest non-versioned baseline for all three backbones (e.g., 0.25 vs.\ 0.21 on Llama-3.1-8B). Prompt-only rollback performs poorly, suggesting that instructing the model to ``ignore'' post-$v^*$ information is insufficient for counterfactual summarization once the agent has been exposed to subsequent interactions. Retrieval-only rollback can outperform full-history prompting on LoCoMo, consistent with the intuition that providing the entire contaminated history makes it harder for the model to maintain a clean temporal scope. On MAB En.Sum, \name\ again yields the strongest F1 across backbones; here, full-history prompting outperforms retrieval-only rollback, reflecting that summarization in this setting depends on global coherence and cannot be reliably reconstructed from a small set of retrieved snippets. Overall, these results confirm that rollback-consistent summarization benefits from \emph{global} memory restoration and state isolation rather than from prompt-based suppression or snippet-level retrieval alone.

\begin{table}[h]
\centering
\small
\begin{tabular}{l|ccc}
\toprule
\textbf{Method} 
& \textbf{Llama-3.1-8B} 
& \textbf{Qwen2.5-7B} 
& \textbf{Mistral-7B} \\
\midrule
Prompt-only & 0.10 & 0.09 & 0.04 \\
Full-history & 0.17 & 0.13 &  0.11 \\
RAG-only & 0.21 & 0.20 & 0.14 \\
\name & \textbf{0.25} & \textbf{0.22} & \textbf{0.17} \\
\bottomrule
\end{tabular}
\caption{Rollback-consistent summarization task ROUGE-1 reported in [0,1] score across different base models on LoCoMo dataset. }
\label{tab:rq3_locomo}
\end{table}

\begin{table}[h]
\centering
\small
\begin{tabular}{l|ccc}
\toprule
\textbf{Method} 
& \textbf{Llama-3.1-8B} 
& \textbf{Qwen2.5-7B} 
& \textbf{Mistral-7B} \\
\midrule
Prompt-only   & 2.0\%  & 3.0\%  & 0.0\% \\
Full-history  & 8.0\%  & 8.0\%  & 1.0\% \\
RAG-only      & 3.0\%  & 2.0\%  & 0.0\% \\
\name         & \textbf{13.0\%} & \textbf{14.0\%} & \textbf{7.0\%} \\
\bottomrule
\end{tabular}
\caption{Rollback-consistent summarization on MAB (En.Sum), reported as F1 (\%).}
\label{tab:rq3_mab}
\end{table}



\section{Conclusion}
We presented \name, an open-source semantic version-control layer for agent memory integrated into a production-ready agent framework (ADK). \name treats long-term agent memory as a versioned state machine: each memory write produces a whole-memory snapshot (a commit), and rollback restores a prior global snapshot so that subsequent reads are scoped to the restored version (\texttt{HEAD}). This design reduces natural-language ``undo'' to a systems problem of \emph{NL$\rightarrow$version resolution} followed by a deterministic restore primitive, rather than relying on prompting or best-effort retrieval to suppress later information. Across two long-horizon benchmarks adapted to a post-exposure rollback setting, \name{} achieves the strongest semantic version selection performance and yields consistent downstream improvements after rollback. Compared to the strongest baseline without global snapshot restoration, \name{} improves rollback-consistent QA and summarization by approximately 10 percentage points on average across datasets and backbone models. These results support a central takeaway: rollback-consistent behavior under post-exposure is primarily an \emph{architectural property} enabled by explicit version-scoped state restoration, not something that emerges reliably from instruction following or snippet-level similarity search alone. We believe \name{} provides a practical foundation for controllable, auditable agent memory in open-source agent stacks. Promising future directions include rollback-native benchmarks with more realistic user rollback intents, branching histories beyond linear truncation, and higher-concurrency memory backends with stronger multi-writer transactional semantics.

\section*{Limitations}

\paragraph{Concurrency and transactional semantics.}
\name\ serializes commit/rollback operations with a per-user lock and relies on SQLite transactions for atomic local state transitions.
However, we do not experimentally evaluate high-concurrency settings (e.g., many simultaneous sessions/users issuing frequent commits and rollbacks).
This is particularly relevant because SQLite serializes writes (single-writer at a time) and uses file-level locking; write-heavy workloads may experience contention and queueing.
A rigorous evaluation of concurrency, contention, and multi-writer scalability (e.g., WAL/WAL2 optimizations) remains future work.

\paragraph{Linear history only (no branching/merging).}
Our current implementation maintains a linear version history and truncates versions after a rollback (Git ``reset''-style).
Supporting branching timelines and merges is conceptually straightforward but is not implemented or evaluated in this work.

\paragraph{Benchmark adaptation and query generation.}
We adapt existing long-horizon memory benchmarks to a post-exposure rollback setting and construct rollback queries using ground-truth-backed anchors with constrained LLM paraphrasing.
While this enables a controlled and systematic evaluation, these benchmarks were not originally designed for versioned rollback scenarios.
Consequently, the induced rollback query distribution may not fully capture the diversity, ambiguity, and partial specification of real-world user rollback intents.

Our results therefore demonstrate the architectural soundness and robustness of semantic rollback under controlled long-horizon settings, rather than claiming a complete characterization of rollback behavior in production-scale agent deployments.


\begin{acks}
This work is supported by the NSF award 2229876.
\end{acks}

\bibliographystyle{ACM-Reference-Format}
\bibliography{main}
\appendix



\end{document}